%% file: main.tex
\documentclass[10pt,twocolumn,letterpaper]{article}

\usepackage{cvpr}
\usepackage{times}
\usepackage{epsfig}
\usepackage{graphicx}
\usepackage{amsmath}
\usepackage{amssymb}
\usepackage{xfrac}

\usepackage{subfigure}
\usepackage[usenames, dvipsnames]{color}
\usepackage{colortbl}

\input{macros}

\usepackage[pagebackref=true,breaklinks=true,letterpaper=true,bookmarks=false]{hyperref}

\cvprfinalcopy 

\def\cvprPaperID{2579} 

\ifcvprfinal\fi
\begin{document}

\title{\papertitle}

\author{
\begin{tabular}{c@{\hspace{0.4in}}c@{\hspace{0.4in}}c}
Tim Brooks$^1$ & Ben Mildenhall$^2$ & Tianfan Xue$^1$ \\
Jiawen Chen$^1$ &  Dillon Sharlet$^1$ &  Jonathan T. Barron$^1$
\end{tabular} \\
 $^1$Google Research, \hspace{0.3in} $^2$UC Berkeley
}

\maketitle

\begin{abstract}

Machine learning techniques work best when the data used for training resembles the data used for evaluation.
This holds true for learned single-image denoising algorithms, which are applied to real raw camera sensor readings but, due to practical constraints, are often trained on synthetic image data.
Though it is understood that generalizing from synthetic to real images requires careful consideration of the noise properties of camera sensors, the other aspects of an image processing pipeline (such as gain, color correction, and tone mapping) are often overlooked, despite their significant effect on how raw measurements are transformed into finished images.
To address this, we present a technique to ``unprocess'' images by inverting
each step of an image processing pipeline, thereby allowing us to synthesize realistic raw sensor measurements from commonly available Internet photos.
We additionally model the relevant components of an image processing pipeline when evaluating our loss function, which allows training to be aware of all relevant photometric processing that will occur after denoising.
By unprocessing and processing training data and model outputs in this way, we are able to train a simple
convolutional neural network that has 14$\%$-38$\%$ lower error rates and is 9$\times$-18$\times$ faster
than the previous state of the art on the Darmstadt Noise Dataset~\cite{plotz2017cvpr},
and generalizes to sensors outside of that dataset as well.
\end{abstract}

\section{Introduction}

\newcommand{\teaserresultswidth}{0.472\linewidth}
\begin{figure}[t]
\begin{center}
  \subfigure[Noisy Input,  PSNR = 18.76]{
  \includegraphics[width=\teaserresultswidth]{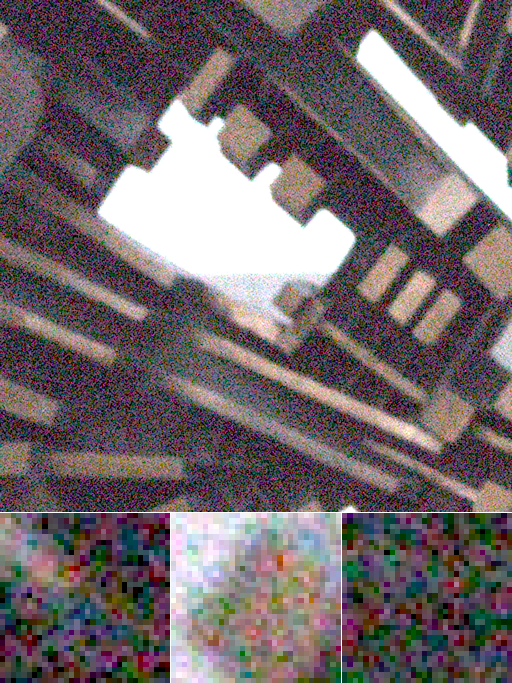}
  \label{subfig:teaser2}
  }
  \subfigure[Ground Truth]{
  \includegraphics[width=\teaserresultswidth]{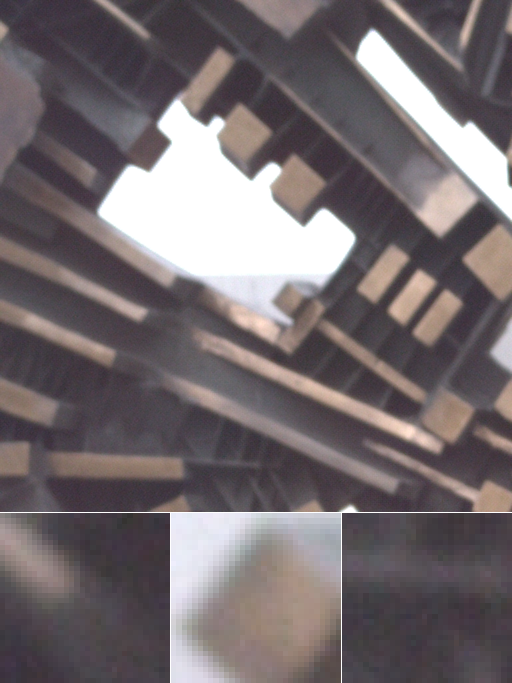}
  \label{subfig:teaser1}
  }
  \subfigure[N3Net \cite{NNN}, PSNR = 32.42]{
  \includegraphics[width=\teaserresultswidth]{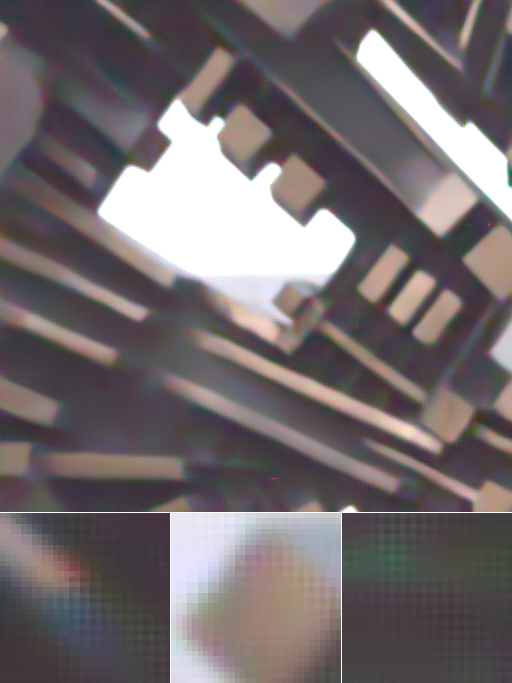}
  \label{subfig:teaser3}
  }
  \subfigure[Our Model, PSNR = 35.35]{
  \includegraphics[width=\teaserresultswidth]{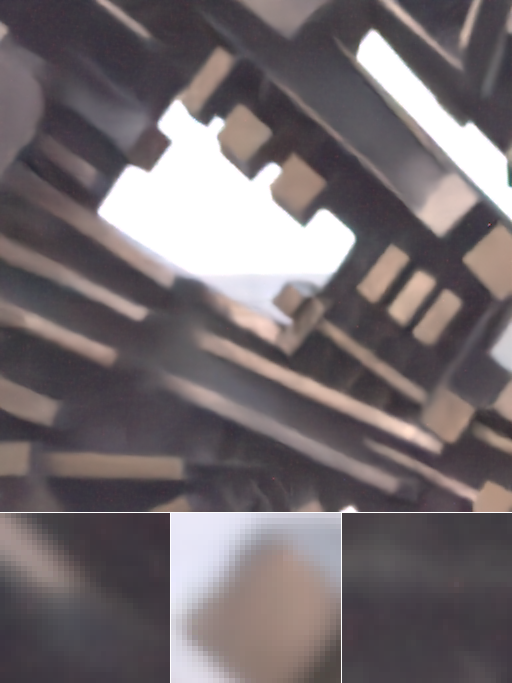}
  \label{subfig:teaser4}
  }
\end{center}
  \caption{An image from the Darmstadt Noise Dataset~\cite{plotz2017cvpr}, where we present \subref{subfig:teaser2}~the noisy input image, \subref{subfig:teaser1}~the ground truth noise-free image, \subref{subfig:teaser3}~the output of the previous state-of-the-art algorithm, and \subref{subfig:teaser4}~the output of our model. All four images were converted from raw Bayer space to sRGB for visualization.
  Alongside each result are three cropped sub-images, rendered with nearest-neighbor interpolation.
  See the supplement for additional results.}
\label{fig:teaser}
\end{figure}

Traditional single-image denoising algorithms often analytically model properties of images and the noise they are designed to remove. In contrast, modern denoising methods often employ neural networks to learn a mapping from noisy images to noise-free images. Deep learning is capable of representing complex properties of images and noise, but training these models requires large paired datasets. As a result, most learning-based denoising techniques rely on synthetic training data. Despite significant work on designing neural networks for denoising, recent benchmarks \cite{anaya2014renoir,plotz2017cvpr} reveal that deep learning models are often outperformed by traditional, hand-engineered algorithms when evaluated on real noisy raw images.

We propose that this discrepancy is in part due to unrealistic synthetic training data.
Many classic algorithms generalize poorly to real data due to assumptions that noise is additive, white, and Gaussian \cite{Rudin1992,Simoncelli1996}.
Recent work has identified this inaccuracy and shifted to more sophisticated noise models that better match the physics of image formation~\cite{Liu2008,Mildenhall_2018_CVPR}. However, these techniques do not consider the many steps of a typical image processing pipeline.

One approach to ameliorate the mismatch between synthetic training data and real raw images is to capture noisy and noise-free image pairs using the same camera being targeted by the denoising algorithm~\cite{SIDD, chen2018cvpr, schwartz2018itip}. However, capturing noisy and noise-free image pairs is difficult, requiring long exposures or large bursts of images, and post-processing to combat camera motion and lighting changes. Acquiring these image pairs is expensive and time consuming, a problem that is exacerbated by the large amounts of training data required to prevent over-fitting when training neural networks. Furthermore, because different camera sensors exhibit different noise characteristics, adapting a learned denoising algorithm to a new camera sensor may require capturing a new dataset.

When properly modeled, synthetic data is simple and effective. The physics of digital sensors and the steps of an imaging pipeline are well-understood and can be leveraged to generate training data from almost any image using only basic information about the target camera sensor. We present a systematic approach for modeling key components of image processing pipelines, ``unprocessing'' generic Internet images to produce realistic raw data, and integrating conventional image processing operations into the training of a neural network. 
When evaluated on real noisy raw images in the Darmstadt Noise Dataset~\cite{plotz2017cvpr}, our model has 14$\%$-38$\%$ lower error rates and is 9$\times$-18$\times$ faster than the previous state of the art. A visualization of our model's output can be seen in Figure~\ref{fig:teaser}. Our unprocessing and processing approach also generalizes images captured from devices which were not explicitly modeled when generating our synthetic training data.

This paper proceeds as follows: In Section~\ref{sec:related} we review related work.
In Section~\ref{sec:image_formation_pipeline} we detail the steps of a raw image processing pipeline and define the inverse of each step.
In Section~\ref{sec:model} we present procedures for unprocessing generic Internet images into synthetic raw data, modifying training loss to account for raw processing, and training our simple and effective denoising neural network model.
In Section~\ref{sec:results} we demonstrate our model's improved performance on the Darmstadt Noise Dataset~\cite{plotz2017cvpr} and provide an ablation study isolating the relative importance of each aspect of our approach.

\section{Related Work}
\label{sec:related}

Single image denoising has been the focus of a significant body of research in computer vision and image processing.
Classic techniques such as anisotropic diffusion~\cite{PeronaMalik1990}, total variation denoising~\cite{Rudin1992}, and wavelet coring~\cite{Simoncelli1996}
use hand-engineered algorithms to recover a clean signal from noisy input, under the assumption that both signal and noise exhibit particular statistical regularities.
Though simple and effective,
these parametric models are limited in their capacity and expressiveness, which led to
increased interest in nonparametric, self-similarity-driven techniques such as BM3D~\cite{BM3D} and non-local means~\cite{NonlocalMeans}.
The move from simple, analytical techniques towards data-driven approaches continued in the form of dictionary-learning and basis-pursuit algorithms such as KSVD~\cite{KSVD} and Fields-of-Experts~\cite{FoE}, which operate by finding image representations where sparsity holds or statistical regularities are well-modeled.
In the modern era, most single-image denoising algorithms are entirely data-driven, consisting of deep neural networks trained to regress from noisy images to denoised images \cite{GharbiDemosaic, CBDnet, NNN, Schmidt2014, TWSC, Zhang2017}.

\begin{figure*}[t!]
\begin{center}
   \includegraphics[width=0.99\linewidth]{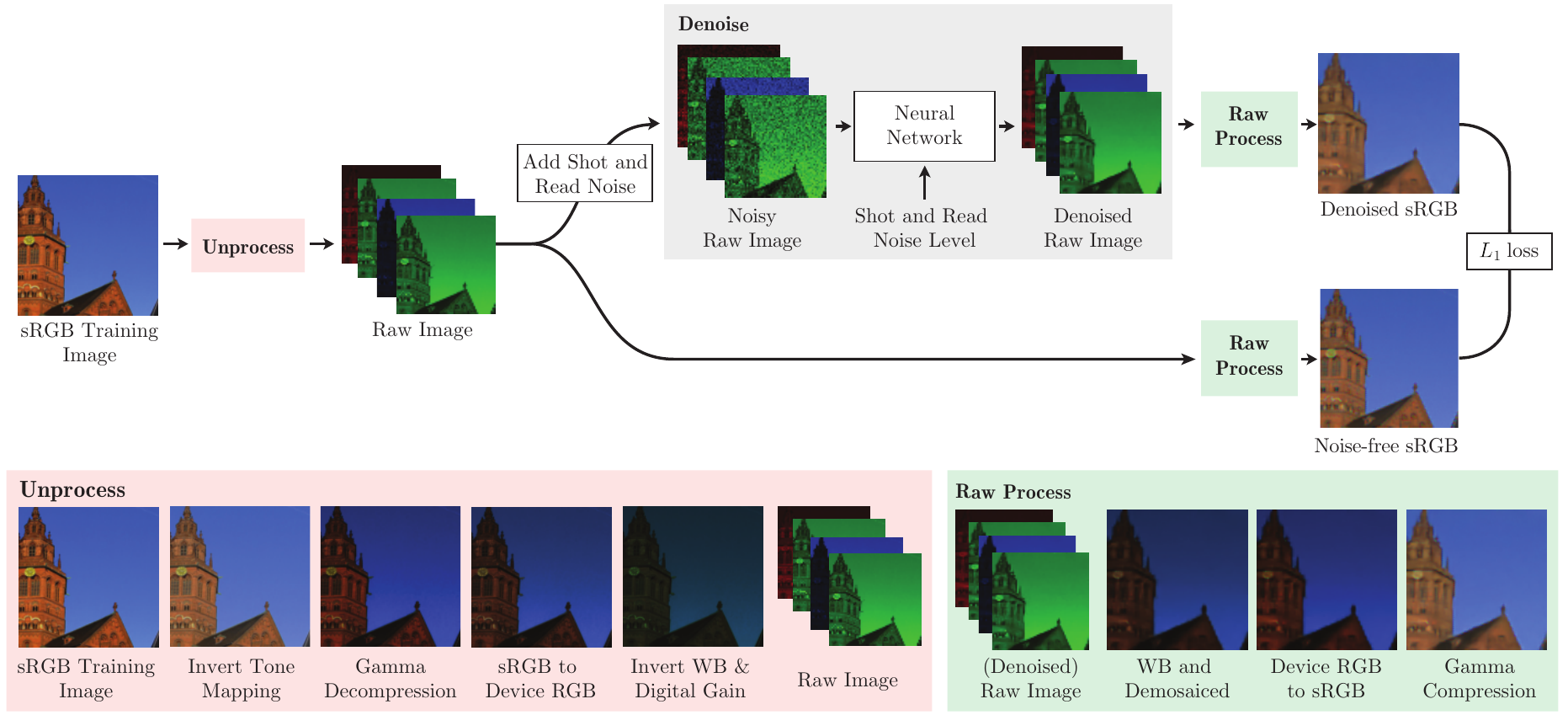}
\end{center}
   \caption{A visualization of our data pipeline and network training procedure. sRGB images from the MIR Flickr dataset~\cite{huiskes10mir} are unprocessed, and realistic shot and read noise is added to synthesize noisy raw input images. Noisy images are fed through our denoising neural network, and the outputs of that network and the noise-free raw images then undergo raw processing before $L_{1}$ loss is computed. See Sections~\ref{sec:image_formation_pipeline} and \ref{sec:model} for details.}
\label{fig:datapipeline}
\end{figure*}

Most classic denoising work was done under the assumption that image noise is additive, white, and Gaussian.
Though convenient and simple, this model is not realistic, as the stochastic process of photons arriving at a sensor is better
modeled as ``shot'' and ``read'' noise~\cite{Hasinoff2014}.
The overall noise can more accurately be modeled as containing both Gaussian and Poissonian signal-dependent components \cite{PoissonianGaussianNoise} or as being sampled from a heteroscedastic Gaussian where variance is a function of intensity \cite{Hasinoff2010}.
An alternative to analytically modeling image noise is to use examples of real noisy and noise-free images.
This can be done by capturing datasets consisting of pairs of real photos, where one image is a short exposure and therefore noisy, and the other image is a long exposure and therefore largely noise-free \cite{anaya2014renoir,plotz2017cvpr}.
These datasets enabled the observation that recent learned techniques trained using synthetic data were outperformed by older models, such as BM3D~\cite{anaya2014renoir,plotz2017cvpr}. 
As a result, recent work has demonstrated progress by collecting this real, paired data not just for evaluation, but for training models \cite{SIDD,chen2018cvpr,schwartz2018itip}. These approaches show great promise, but applying such a technique to a particular camera requires the laborious collection of large amounts of perfectly-aligned training data for that camera, significantly increasing the burden on the practitioner compared to the older techniques that required only synthetic training data or calibrated parameters. Additionally, it is not clear how this dataset acquisition procedure could be used to capture subjects where small motions are pervasive, such as water, clouds, foliage, or living creatures.
Recent work suggests that multiple noisy images of the same scene can be used as training data instead of paired noisy and noise-free images~\cite{Noise2Noise}, but this does not substantially mitigate the limitations or the labor requirements of these large datasets of real photographs.

Though it is generally understood that correctly modeling noise during image formation is critical for learning an effective denoising algorithm~\cite{Hasinoff2010, Liu2008, Mildenhall_2018_CVPR, NNN}, a less well-explored issue is the effect of the image processing pipeline used to turn raw sensor readings into a finished image.
Modern image processing pipelines (well described in \cite{hasinoff2016burst}) consist of several steps which transform image intensities, therefore effecting both how input noise is scaled or modified and how the final rendered image appears as a function of the raw sensor measurements.
In this work we model and invert these same steps when synthesizing training data for our model, and demonstrate that doing so significantly improves denoising performance.

\section{Raw Image Pipeline}
\label{sec:image_formation_pipeline}

Modern digital cameras attempt to render a pleasant and accurate image of the world, similar to that perceived by the human eye. However, the raw sensor data from a camera does not yet resemble a photograph, and many processing stages are required to transform its noisy linear intensities into their final form. In this section, we describe a conventional image processing pipeline, proceeding from sensor measurement to a final image. To enable the generation of realistic synthetic raw data, we also describe how each step in our pipeline can be inverted. Through this procedure we are able to turn generic Internet images into training pairs that well-approximate the Darmstadt Noise Dataset~\cite{plotz2017cvpr}, and generalize well to other raw images.
See Figure~\ref{fig:datapipeline} for an overview of our unprocessing steps.

\subsection{Shot and Read Noise}
\label{sec:noise}

Though the noise in a processed image may have very complex characteristics due to nonlinearities and correlation across pixel values, the noise in raw sensor data is well understood.
Sensor noise primarily comes from two sources: photon arrival statistics (``shot'' noise) and imprecision in the readout circuitry (``read'' noise)~\cite{Hasinoff2014}. Shot noise is a Poisson random variable whose mean is the true light intensity (measured in photoelectrons). Read noise is an approximately Gaussian random variable with zero mean and fixed variance. We can approximate these together as a single heteroscedastic Gaussian and treat each observed intensity $y$ as a random variable whose variance is a function of the true signal $x$:
\begin{equation}
y \sim \mathcal N(\mu=x, \sigma^2=\readnoise + \shotnoise x).
\end{equation}
Parameters $\readnoise$ and $\shotnoise$ are determined by sensor's analog and digital gains. For some digital gain $g_d$, analog gain $g_a$, and fixed sensor readout variance $\sigma_r^2$, we have
\begin{equation}
\readnoise = g_d^2 \sigma_r^2,\quad \shotnoise = g_d g_a.
\end{equation}
These two gain levels are set by the camera as a direct function of the ISO light sensitivity level chosen by the user or by some auto exposure algorithm. Thus the values of $\readnoise$ and $\shotnoise$ can be calculated by the camera for a particular exposure and are usually stored as part of the metadata accompanying a raw image file.

\begin{figure}[t]
\begin{center}
   \includegraphics[width=0.99\linewidth]{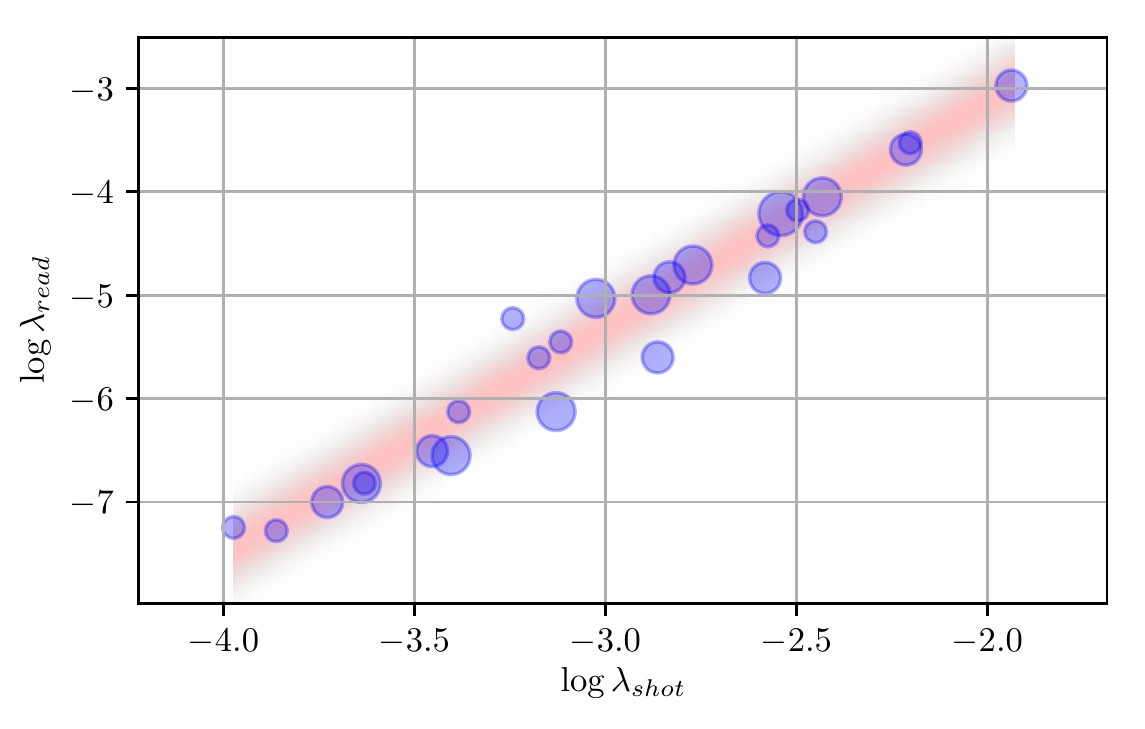}
\end{center}
   \caption{Shot and read noise parameters from the Darmstadt dataset~\cite{plotz2017cvpr}. The size of each circle indicates how many images in the dataset shared that shot/read noise pair. To choose the noise level for each synthetic training image, we randomly sample shot and read noise parameters from the distribution shown in red.}
\label{fig:noiseparams}
\end{figure}

To choose noise levels for our synthetic images, we model the joint distribution of different shot/read noise parameter pairs in our real raw images and sample from that distribution.
For the Darmstadt Noise Dataset~\cite{plotz2017cvpr}, a reasonable sampling procedure of shot/read noise factors is
\begin{align}
&\logshotnoise \sim {\mathcal {U}}(a=\log(0.0001),b=\log(0.012)) \nonumber \\
&\logreadnoise \mid \logshotnoise \sim \nonumber \\
& \quad\quad\quad\quad {\displaystyle {\mathcal {N}}(\mu=2.18 \logshotnoise+ 1.2 ,\sigma=0.26)} .
\end{align}
See Figure~\ref{fig:noiseparams} for a visualization of this process.

\subsection{Demosaicing}
\label{sec:demosaic}

Each pixel in a conventional camera sensor is covered by a single red, green, or blue color filter, arranged in a Bayer pattern, such as R-G-G-B. The process of recovering all three color measurements for each pixel in the image is the well-studied problem of demosaicing~\cite{GharbiDemosaic}. The Darmstadt dataset follows the convention of using bilinear interpolation to perform demosaicing, which we adopt. Inverting this step is trivial---for each pixel in the image we omit two of its three color values according to the Bayer filter pattern.

\subsection{Digital Gain}
\label{sec:gain}

A camera will commonly apply a digital gain to all image intensities, where each image's particular gain is selected by the camera's auto exposure algorithm. These auto exposure algorithms are usually proprietary ``black boxes'' and are difficult to reverse engineer for any individual image. But to invert this step for a pair of synthetic and real datasets, a reasonable heuristic is to simply find a single global scaling that best matches the marginal statistics of all image intensities across both datasets.
To produce this scaling, we assume that our real and synthetic image intensities are both drawn from different exponential distributions:
\begin{equation}
\displaystyle p(x ; \lambda )=\lambda e^{-\lambda x}
\end{equation}
for $x \geq 0$. The maximum likelihood estimate of the scale parameter $\lambda$ is simply the inverse of the sample mean, and scaling $x$ is equivalent to an inverse scaling of $\lambda$. This means that we can match two sets of intensities that are both exponentially distributed by using the ratio of the sample means of both sets. When using our synthetic data
and the Darmstadt dataset, this scaling ratio is $1.25$. For more thorough data augmentation and to ensure that our model observes pixel intensities throughout $[0, 1]$ during training, rather than applying this constant scaling, we sample inverse gains from a normal distribution centered at $1 / 1.25 = 0.8$ with standard deviation of $0.1$, resulting in inverse gains roughly spanning $[0.5, 1.1]$.

\subsection{White Balance}
\label{sec:wb}

The image recorded by a camera is the product of the color of the lights that illuminate the scene and the material colors of the objects in the scene.
One goal of a camera pipeline is to undo some of the effect of illumination, producing an image that appears to be lit under ``neutral'' illumination.
This is performed by a white balance algorithm that estimates a per-channel gain for the red and blue channels of an image using a heuristic or statistical approach \cite{Gijsenij2011,BarronTsai2017}.
Inverting this procedure from synthetic data is challenging because, like auto exposure, the white balance algorithm of a camera is unknown and therefore difficult to reverse engineer. However, raw image datasets such as Darmstadt record the white balance metadata of their images, so we can synthesize somewhat realistic data by simply sampling from the empirical distribution of white balance gains in that dataset: a red gain in $[1.9, 2.4]$ and a blue gain in $[1.5, 1.9]$, sampled uniformly and independently.

\begin{figure}[t]
\begin{center}
   \includegraphics[width=0.5\linewidth]{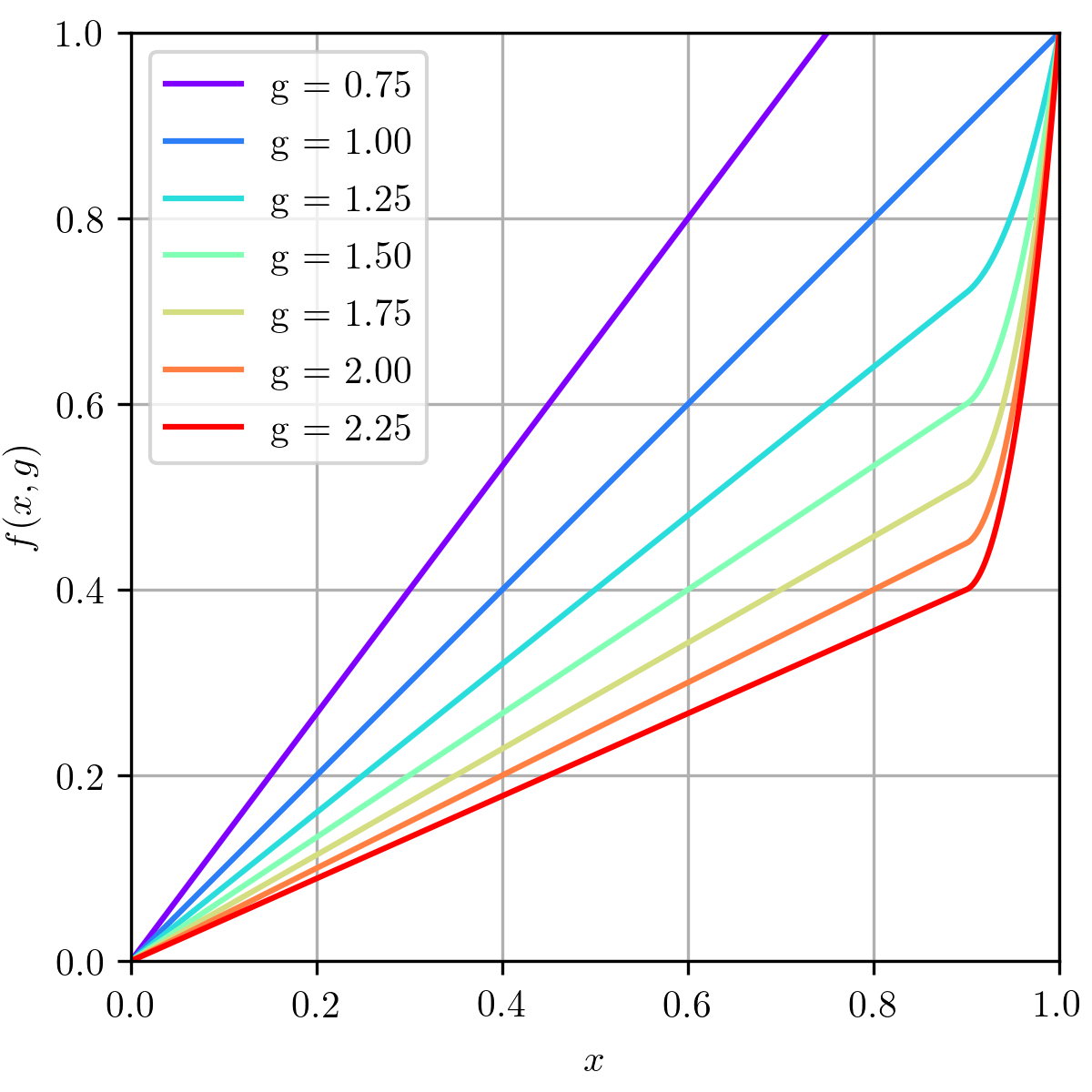}
\end{center}
   \caption{The function $f(x, g)$ (defined in Equation~\ref{equ:gainfn}) we use for gaining down synthetic image intensities $x$ while preserving highlights, for a representative set of gains $\{g\}$.}
\label{fig:dim_curve}
\end{figure}

When synthesizing training data, we sample inverse digital and white balance gains and take their product to get a per-channel inverse gain to apply to our synthetic data.
This inverse gain is almost always less than unity, which means that na\"ively gaining down our synthetic imagery will result in a dataset that systematically lacks highlights and contains almost no clipped pixels. This is problematic, as correctly handling saturated image intensities is critical when denoising. To account for this, instead of applying our inverse gain $\sfrac{1}{g}$ to some intensity $x$ with a simple multiplication, we apply a highlight-preserving transformation $f(x,g)$ that is linear when $g \leq 1$ or $x \leq t$ for some threshold $t=0.9$, but is a cubic transformation when $g > 1$ and $x > t$:
\begin{align}
    \alpha(x) &= \left(\frac{\max(x - t, 0)}{1-t}\right)^2
    \\
    f(x,g) &= \max\left(\frac{x}{g},\,\, \left(1-\alpha(x)\right) \left( \frac{x}{g} \right) + \alpha(x) x\right)
    \label{equ:gainfn}
\end{align}
This transformation is designed such that $f(x,g) = \sfrac{x}{g}$ when $x \leq t$, $f(1,g) = 1$ when $g \leq 1$, and $f(x, g)$ is continuous and differentiable. This function is visualized in Figure~\ref{fig:dim_curve}.

\subsection{Color Correction}
\label{sec:ccm}

In general, the color filters of a camera sensor do not match the spectra expected by the sRGB color space. To address this, a camera will apply a $3\times3$ color correction matrix (CCM) to convert its own ``camera space'' RGB color measurements to sRGB values. The Darmstadt dataset consists of four cameras, each of which uses its own fixed CCM when performing color correction. To generate our synthetic data such that it will generalize to all cameras in the dataset, we sample random convex combinations of these four CCMs, and for each synthetic image, we apply the inverse of a sampled CCM to undo the effect of color correction.

\subsection{Gamma Compression}
\label{sec:gamma}

Because humans are more sensitive to gradations in the dark areas of images, gamma compression is typically used to allocate more bits of dynamic range to low intensity pixels. We use the same standard gamma curve as \cite{plotz2017cvpr}, while taking care to clamp the input to the gamma curve with $\epsilon=10^{-8}$ to prevent numerical instability during training:
\begin{equation}
\Gamma(x) = \max(x, \epsilon)^{\sfrac{1}{2.2}}
\end{equation}
When generating synthetic data, we apply the (slightly approximate, due to $\epsilon$) inverse of this operator:
\begin{equation}
\Gamma^{-1}(y) = \max(y, \epsilon)^{2.2}
\end{equation}

\subsection{Tone Mapping}
\label{sec:tonemap}
While high dynamic range images require extreme tone mapping~\cite{Debevec1997}, even standard low-dynamic-range images are often processed with an S-shaped curve designed to match the ``characteristic curve'' of film \cite{davis1922sensitometry}.
More complex edge-aware local tone mapping may be performed, though reverse-engineering such an operation is difficult \cite{Paris2011}.
We therefore assume that tone mapping is performed with a simple ``smoothstep'' curve, and we use the inverse of that curve when generating synthetic data.
\begin{align}
\displaystyle \operatorname {smoothstep}(x)&= 3x^{2}-2x^{3}\\
\displaystyle \operatorname {smoothstep}^{-1}(y)&= \frac{1}{2} - {\sin}\left(\frac{{\sin}^{-1}(1-2y) }{ 3}\right)
\end{align}
where both are only defined on inputs in $[0, 1]$.

\section{Model}
\label{sec:model}

Now that we have defined each step of our image processing pipeline and each step's inverse, we can construct our denoising neural network model. The input and ground-truth used to train our network is synthetic data that has been unprocessed using the inverse of our image processing pipeline, where the input image has additionally been corrupted by noise.
The output of our network and the ground-truth are processed by our pipeline before evaluating the loss being minimized.

\newcommand{\histwidth}{0.305\linewidth}
\begin{figure}[t]
\begin{center}
  \subfigure[sRGB]{
  \label{subfig:hist1}
  \includegraphics[width=\histwidth]{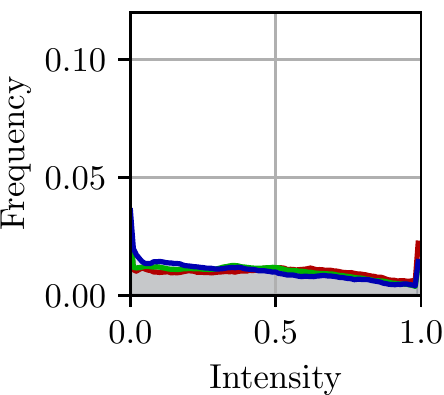}
  }
  \subfigure[Unprocessed]{
  \label{subfig:hist2}
  \includegraphics[width=\histwidth]{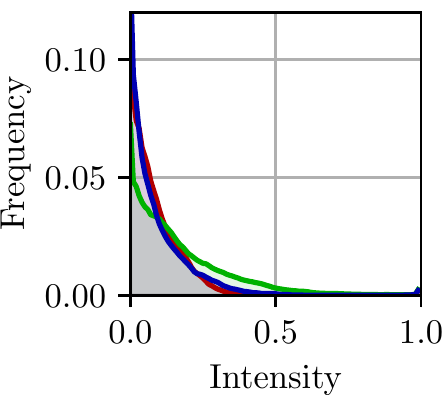}
  }
  \subfigure[Raw]{
  \label{subfig:hist0}
  \includegraphics[width=\histwidth]{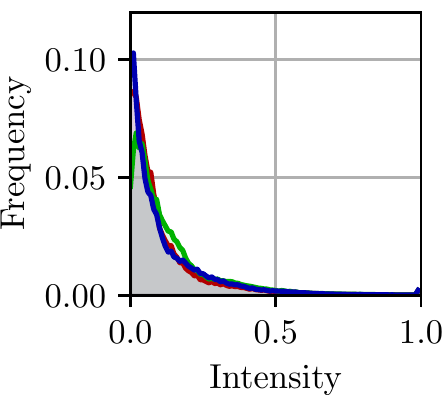}
  }
\end{center}
   \caption{
   Histograms for each color channel of \subref{subfig:hist1} sRGB images from the MIR Flickr dataset, \subref{subfig:hist2}~unprocessed images created following the procedure enumerated in Section~\ref{sec:preproc} and detailed in Section~\ref{sec:image_formation_pipeline}, and \subref{subfig:hist0}~real raw images from the Darmstadt dataset. Note that the distributions of real raw intensities and our unprocessed intensities are similar.
}
   \label{fig:histograms}
\end{figure}

\subsection{Unprocessing Training Images}
\label{sec:preproc}

To generate realistic synthetic raw data, we unprocess images by sequentially inverting image processing transformations, as summarized in Figure~\ref{fig:datapipeline}. This consists of inverting, in order, tone mapping (Section~\ref{sec:tonemap}), applying gamma decompression (Section~\ref{sec:gamma}), applying the sRGB to camera RGB color correction matrix (Section~\ref{sec:ccm}), and inverting white balance gains (Section~\ref{sec:wb}) and digital gain (Section~\ref{sec:gain}). The resulting synthetic raw image is used as the noise-free ground truth during training, and shot and read noise (Section~\ref{sec:noise}) is added to create the noisy network input. Our synthetic raw images more closely resemble real raw intensities, as demonstrated in Figure~\ref{fig:histograms}.

\subsection{Processing Raw Images}
\label{sec:postproc}

Since raw images ultimately go through an image processing pipeline before being viewed, the output images from our model should also be subject to such a pipeline before any loss is evaluated. We therefore apply raw processing to the output of our model, which in order consists of applying white balance gains (Section~\ref{sec:wb}), na\"ive bilinear demosaicing (Section~\ref{sec:demosaic}), applying a color correction matrix to convert from camera RGB to sRGB (Section~\ref{sec:ccm}), and gamma compression (Section~\ref{sec:gamma}). This simplified image processing pipeline matches that used in the Darmstadt Noise Dataset benchmark~\cite{plotz2017cvpr} and is a good approximation for general image pipelines. We apply this processing to the network's output and to the ground truth noise-free image before computing our loss. Incorporating this pipeline into training allows the network to reason about how downstream processing will impact the desired denoising behavior.

\begin{figure}[t]
\begin{center}
   \includegraphics[width=0.99\linewidth]{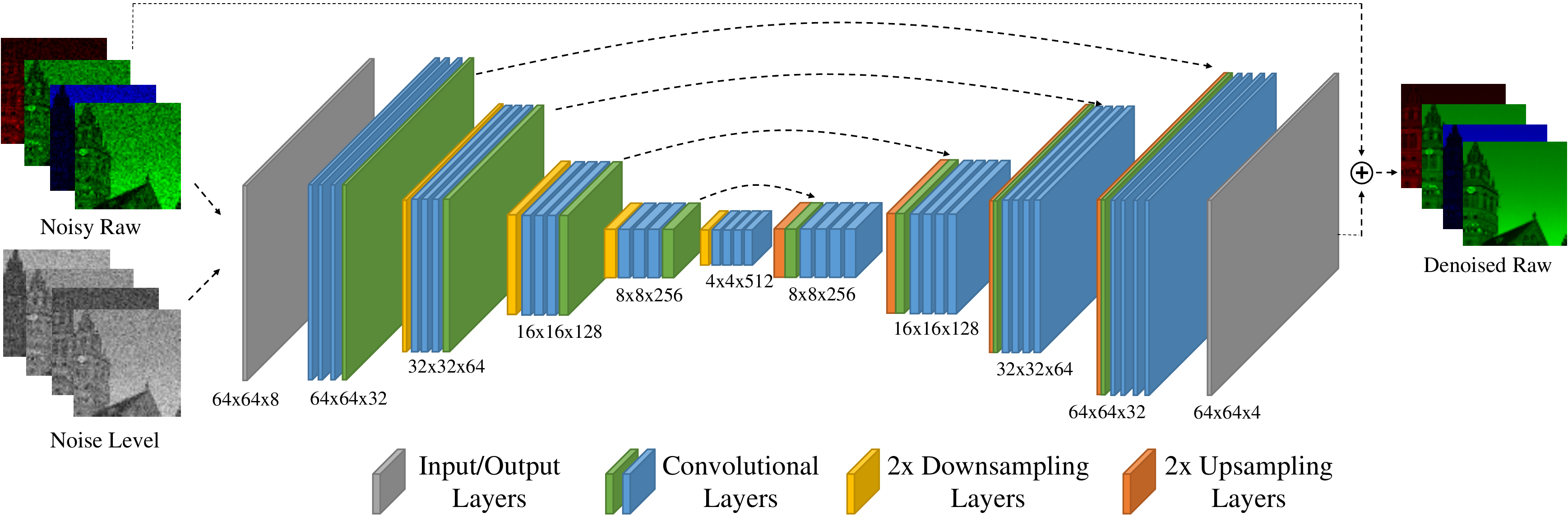}
\end{center}
   \caption{The network structure of our model. Input to the network is a 4-channel noisy mosaic image concatenated with a 4-channel noise level map, and output is a 4-channel denoised mosaic image.
}
\label{fig:network}
\end{figure}

\subsection{Architecture}
\label{sec:arch}

Our denoising network takes as input a noisy raw image in the Bayer domain and outputs a reduced noise image in the same domain. As an additional input, we pass the network a per-pixel estimate of the standard deviation of noise in the input image, based on its shot and read noise parameters. This information is concatenated to the input as $4$ additional channels---one for each of the R-G-G-B Bayer planes. We use a U-Net architecture~\cite{unet} with skip connections between encoder and decoder blocks at the same scale (see~Figure~\ref{fig:network} for details), with box downsampling when encoding, bilinear upsampling when decoding, and the PReLU~\cite{he2015delving} activation function. 
As in~\cite{Zhang2017}, instead of directly predicting a denoised image, our model predicts a residual that is added back to the input image.

\subsection{Training}
\label{sec:training}

To create our synthetic training data, we start with the 1 million images of the MIR Flickr extended dataset~\cite{huiskes10mir}, setting aside $5\%$ of the dataset for validation and $5\%$ for testing. We downsample all images by 2$\times$ using a Gaussian kernel ($\sigma=1$) to reduce the effect of noise, quantization, JPEG compression, demosaicing, and other artifacts. We then take random $128 \times 128 $ crops of each image, with random horizontal and vertical flips for data augmentation. We synthesize noisy and clean raw training pairs by applying the unprocessing steps described in Section~\ref{sec:preproc}.
We train using Adam \cite{KingmaB14} with a learning rate of $10^{-4}$, $\beta_1=0.9$, $\beta_2=0.999$, $\epsilon=10^{-7}$, and a batch size of $16$. Our models and ablations are trained to convergence over approximately $3.5$ million steps on a single NVIDIA Tesla P100 GPU, which takes $\sim$3 days.

We train two models, one targeting performance on sRGB error metrics, and another targeting performance on raw error metrics. For our ``sRGB'' model the network output and synthetic ground-truth are both transformed to sRGB space before computing the loss, as described in Section~\ref{sec:postproc}. Our ``Raw'' model instead computes the loss directly between our network output and our raw synthetic ground-truth, without this processing. For both experiments we minimize $L_1$ loss between the output and ground-truth images.

\definecolor{Yellow}{rgb}{1,1, 0.6}
\definecolor{Orange}{rgb}{1,0.8, 0.6}
\definecolor{Red}{rgb}{1, 0.6, 0.6}

\section{Results}
\label{sec:results}

\begin{table*}[]
\begin{center}
\resizebox{\linewidth}{!}{
\begin{tabular}{ l || cc|cc || cc|cc || c }
\multicolumn{1}{c||}{} & 
\multicolumn{4}{c||}{Raw} & 
\multicolumn{4}{c||}{sRGB}& Runtime \\
Algorithm & \multicolumn{2}{c|}{PSNR} & \multicolumn{2}{c||}{SSIM} & \multicolumn{2}{c|}{PSNR} & \multicolumn{2}{c||}{SSIM} & (ms) \\
\hline
\input{darmstadt.tex}
\multicolumn{10}{c}{} \\
\multicolumn{10}{c}{Ablations of ``Our Model (sRGB)''} \\
\hline
\input{ablations.tex}
\multicolumn{10}{c}{} \\
\end{tabular}
}
\caption{Performance of our model and its ablations on the Darmstadt Noise Dataset \cite{plotz2017cvpr} compared to all published techniques at the time of submission, taken from \url{https://noise.visinf.tu-darmstadt.de/benchmark/}, and sorted by sRGB PSNR. For baseline methods that have been benchmarked with and without a variance stabilizing transformation (VST), we report whichever version performs better and indicate accordingly in the algorithm name.
We report baseline techniques that use either raw or sRGB data as input, and because this benchmark does not evaluate sRGB-input techniques in terms of raw output, the raw error metrics are missing for those techniques.
For each technique and metric we report relative improvement in parenthesis, which is done by turning PSNR into RMSE and SSIM into DSSIM and then computing the reduction in error relative to the best-performing models.
Ablations of our model are presented in a separate sub-table.
The top three techniques for each metric (ignoring ablations) are color-coded.
Runtimes are presented when available (see Section~\ref{sec:runtimes}).
\label{table:darmstadt}}
\end{center}
\end{table*}

To evaluate our technique we use the Darmstadt Noise Dataset \cite{plotz2017cvpr}, a benchmark of 50 real high-resolution images where each noisy high-ISO image is paired with a (nearly) noise-free low-ISO ground-truth image.
The Darmstadt dataset represents a significant improvement upon earlier benchmarks for denoising, which tended to rely on synthetic data and synthetic (and often unrealistic) noise models.
Additional strengths of the Darmstadt dataset are that it includes images taken from four different standard consumer cameras of natural ``in the wild'' scene content, where the camera metadata has been captured and the camera noise properties have been carefully calibrated, and where the image intensities are presented as raw unprocessed linear intensities.
Another valuable property of this dataset is that evaluation on the dataset is restricted through a carefully controlled online submission system: the entire dataset is the test set, with the ground-truth noise-free images completely hidden from the public, and the frequency of submissions to the dataset is limited. As a result, overfitting to the test set of this benchmark is difficult.
Though this approach is common for object recognition \cite{Everingham2010} and stereo \cite{Middlebury} challenges, it is not common in the context of image denoising.

The performance of our model on the Darmstadt dataset with respect to prior work is shown in Table~\ref{table:darmstadt}. The Darmstadt dataset as presented by \cite{plotz2017cvpr} separates its evaluation into multiple categories: algorithms that do and do not use a variance stabilizing transformation, and algorithms that use linear Bayer sensor readings or that use bilinearly demosaiced sRGB images as input. Each algorithm that operates on raw input is evaluated both on raw Bayer images, and on their denoised Bayer outputs after conversion to sRGB space. Following the procedure of the Darmstadt dataset, we report PSNR and SSIM for each technique, on raw and sRGB outputs. Some algorithms only operate on sRGB inputs; to be as fair as possible to all prior work, we present these models, reporting their evaluation in sRGB space. For algorithms which have been evaluated with and without a variance stabilizing transformation (VST), we include whichever version performs better.

The two variants of our model (one targeting sRGB and the other targeting raw) produce significantly higher PSNRs and SSIMs than all baseline techniques across all outputs, with each model variant outperforming the other for the domain that it targets.
Relative improvements on PSNR and SSIM are difficult to judge, as both metrics are designed to saturate as errors become small. To help with this, alongside each error we report the relative reduction in error of the best-performing model with respect to that model, in parentheses. This was done by converting PSNR into RMSE ($\textrm{RMSE} \propto \sqrt{10^{-\textrm{PSNR}/10}}$) and converting SSIM into DSSIM ($\textrm{DSSIM} = (1 - \textrm{SSIM})/2$) and then computing each relative reduction in error.

We see that our models produce a 14$\%$ and 25$\%$ reduction in error on the two raw metrics compared to the next best performing technique (N3Net~\cite{NNN}), and a 21$\%$ and 38$\%$ reduction in error on the two sRGB metrics compared to the two next best performing techniques (N3Net~\cite{NNN} and CBDNet~\cite{CBDnet}). Visualizations of our model's output compared to other methods can be seen in Figure~\ref{fig:teaser} and in the supplement.
Our model's improved performance appears to be partly due to the decreased low-frequency chroma artifacts in its output compared to our baselines.

\begin{figure}[t]
\begin{center}
  \subfigure[Noisy Input]{
  \includegraphics[width=\teaserresultswidth]{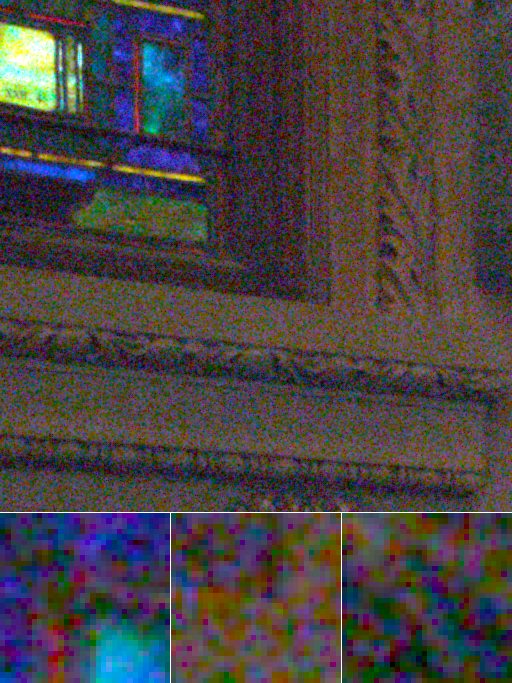}
  \label{subfig:hdrp1}
  }
  \subfigure[Our Model]{
  \includegraphics[width=\teaserresultswidth]{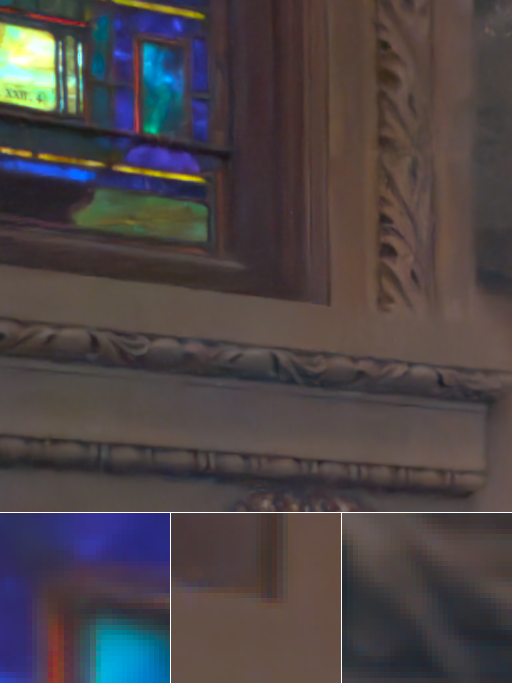}
  \label{subfig:hdrp2}
  }
\end{center}
  \caption{An image from the HDR+ dataset~\cite{hasinoff2016burst}, where we present \subref{subfig:hdrp1}~the noisy input image and \subref{subfig:hdrp2}~the output of our model, in the same format as Figure~\ref{fig:teaser}.
  See the supplement for additional results.}
\label{fig:hdrp}
\end{figure}

To verify that our approach generalizes to other datasets and devices, we evaluated our denoising method on raw images from the HDR+ dataset \cite{hasinoff2016burst}. Results from these evaluations are provided in Figure~\ref{fig:hdrp} and in the supplemental material.

Separately from our two primary models of interest, we present an ablation study of ``Our Model (sRGB),'' in which we remove one or more model components.
``No CCM, WB, Gain'' indicates that when generating synthetic training data we did not perform the unprocessing steps of sRGB to camera RGB CCM inversion, or inverting white balance and digital gain. ``No Tone Mapping, Gamma'' indicates that we did not perform the unprocessing steps of inverting tone mapping or gamma decompression. ``No Unprocessing'' indicates that we did not perform any unprocessing steps, and ``$4\times$ bigger'' indicates that we quadrupled the number of channels in each conv layer. ``Noise-blind'' indicates that the noise level was not provided as input to the network. ``AWGN'' indicates that instead of using our more realistic noise model when synthesizing training data, we use additive white Gaussian noise with $\sigma$ sampled uniformly between $0.001$ and $0.15$ (the range reported in~\cite{plotz2017cvpr}).
``No Residual Output'' indicates that our model architecture directly predicts the output image, instead of predicting a residual that is added to the input.

We see from this ablation study that removing any of our proposed model components reduces quality. Performance is most sensitive to our modeling of noise, as using Gaussian noise significantly decreases performance.
Unprocessing also contributes substantially, especially when evaluated on sRGB metrics, albeit slightly less than a realistic noise model. 
Notably, increasing the network size does not make up for the omission of unprocessing steps.
Our only ablation study that actually removes a component of our neural network architecture (the residual output block) results in the smallest decrease in performance.

\subsection{Runtimes}
\label{sec:runtimes}

Table~\ref{table:darmstadt} also includes runtimes for as many models as we were able to find. Many of these runtimes were produced on different hardware platforms with different timing conventions, so we detail how these numbers were produced here.
The runtime of our model is 22ms for the 512$\times$512 images of the Darmstadt dataset, using our TensorFlow implementation running on a single NVIDIA GeForce GTX 1080Ti GPU, excluding the time taken for data to be transferred to the GPU. We report the mean over 100 runs.
The runtime for DnCNN is taken from \cite{Zhang2017}, which reports a runtime on a GPU (Nvidia Titan X) of 60ms for a 512$\times$512 image, also not including GPU memory transfer times.
The runtime for N3Net \cite{NNN} is taken from that paper, which reports a runtime of 3.5$\times$ that of \cite{Zhang2017}, suggesting a runtime of 210ms.
In \cite{MLP} they report a runtime of 60 seconds on a 512$\times$512 image for a CPU implementation, and note that their runtime is less than that of KSVD \cite{KSVD}, which we note accordingly.
The runtime for CBDNet was taken from \cite{CBDnet}, and the runtimes for BM3D, TNRD, TWSC, and MCWNNM were taken from \cite{TWSC}.
We were unable to find reported runtimes for the remaining techniques in Table~\ref{table:darmstadt}, though in \cite{plotz2017cvpr} they note that  ``many of the benchmarked algorithms are too slow to be applied to megapixel-sized images''.
Our model is the fastest technique by a significant margin: 9$\times$ faster than N3Net~\cite{NNN} and 18$\times$ faster than CBDnet~\cite{CBDnet}, the next two best performing techniques after our own.

\section{Conclusion}

We have presented a technique for ``unprocessing'' generic images into 
data that resembles the raw measurements captured by real camera sensors, by modeling and inverting each step of a camera's image processing pipeline.
This allowed us to train a convolutional neural network for the task of denoising raw image data, where we synthesized large amounts of realistic noisy/clean paired training data from abundantly available Internet images.
Furthermore, by incorporating standard image processing operations into the learning procedure itself, we are able to train a network that is explicitly aware of how its output will be processed before it is evaluated.
When our resulting learned model is applied to the Darmstadt Noise Dataset~\cite{plotz2017cvpr} it achieves 14$\%$-38$\%$ lower error rates and 9$\times$-18$\times$ faster runtimes than the previous state of the art.

{\small
\bibliographystyle{ieee}
\bibliography{references}
}

\end{document}

%% file: macros.tex
\newcommand{\papertitle}{Unprocessing Images for Learned Raw Denoising}

\definecolor{darkred}{rgb}{0.5,0,0}
\definecolor{darkblue}{rgb}{0,0,0.5}
\definecolor{darkgreen}{rgb}{0,0.5,0}
\definecolor{purple}{rgb}{0.5,0,0.5}
\definecolor{yellow}{rgb}{0,0.5,0.5}

\newcommand{\noise}{\lambda}
\newcommand{\shotnoise}{\noise_{\mathit{shot}}}
\newcommand{\logshotnoise}{\log\left(\noise_{\mathit{shot}}\right)}
\newcommand{\logreadnoise}{\log\left(\noise_{\mathit{read}}\right)}
\newcommand{\readnoise}{\noise_{\mathit{read}}}

%% file: darmstadt.tex
FoE \cite{FoE}  &   45.78 &  (30.1\%) &  0.9666 &  (47.3\%) &  35.99 &  (39.5\%) &  0.9042 &  (62.5\%) & - \\
TNRD \cite{TNRD}   + VST &   45.70 &  (30.7\%) &  0.9609 &  (55.0\%) &  36.09 &  (38.8\%) &  0.8883 &  (67.9\%) & 5{,}200 \\
MLP \cite{MLP}   + VST  &   45.71 &  (30.7\%) &  0.9629 &  (52.6\%) &  36.72 &  (34.2\%) &  0.9122 &  (59.1\%) & $\sim$60{,}000 \\
MCWNNM \cite{MCWNNM} &   - &  - &  - &  - &  37.38 &  (29.0\%) &  0.9294 &  (49.2\%) & 208{,}100 \\
EPLL \cite{EPLL}  + VST &   46.86 &  (20.8\%) &  0.9730 &  (34.8\%) &  37.46 &  (28.3\%) &  0.9245 &  (52.5\%) & - \\
KSVD \cite{KSVD}  + VST &   46.87 &  (20.8\%) &  0.9723 &  (36.5\%) &  37.63 &  (26.9\%) &  0.9287 &  (49.6\%) & $>$60{,}000 \\
WNNM \cite{WNNM}  + VST &   47.05 &  (19.1\%) &  0.9722 &  (36.7\%) &  37.69 &  (26.4\%) &  0.9260 &  (51.5\%) & - \\
NCSR \cite{NCSR}  + VST &   47.07 &  (18.9\%) &  0.9688 &  (43.6\%) &  37.79 &  (25.6\%) &  0.9233 &  (53.2\%) & - \\
BM3D \cite{BM3D}  + VST &   47.15 &  (18.2\%) &  0.9737 &  (33.1\%) &  37.86 &  (25.0\%) &  0.9296 &  (49.0\%) & 6{,}900 \\
TWSC \cite{TWSC} &   - &  - &  - &  - &  37.94 &  (24.3\%) &  0.9403 &  (39.9\%) & 195{,}200 \\
CBDNet \cite{CBDnet} &   - &  - &  - &  - &  38.06 &  (23.2\%) & \cellcolor{Yellow} 0.9421 & \cellcolor{Yellow} (38.0\%) & 400 \\
DnCNN \cite{Zhang2017} &   47.37 &  (16.1\%) &  0.9760 &  (26.7\%) &  38.08 &  (23.0\%) &  0.9357 &  (44.2\%) & 60 \\
N3Net \cite{NNN} &  \cellcolor{Yellow} 47.56 & \cellcolor{Yellow} (14.2\%) & \cellcolor{Yellow} 0.9767 & \cellcolor{Yellow} (24.5\%) & \cellcolor{Yellow} 38.32 & \cellcolor{Yellow} (20.9\%) &  0.9384 &  (41.7\%) & 210 \\
Our Model (Raw)     &  \cellcolor{Red} 48.89 & \cellcolor{Red} (0.0\%) & \cellcolor{Red} 0.9824 & \cellcolor{Red} (0.0\%) & \cellcolor{Orange} 40.17 & \cellcolor{Orange} (2.1\%) & \cellcolor{Orange} 0.9623 & \cellcolor{Orange} (4.8\%) & 22 \\
Our Model (sRGB)    &  \cellcolor{Orange} 48.88 & \cellcolor{Orange} (0.1\%) & \cellcolor{Orange} 0.9821 & \cellcolor{Orange} (1.7\%) & \cellcolor{Red} 40.35 & \cellcolor{Red} (0.0\%) & \cellcolor{Red} 0.9641 & \cellcolor{Red} (0.0\%) & 22 \\


%% file: ablations.tex

Noise-blind, AWGN  &   46.48 &  (24.2\%) &  0.9703 &  (40.7\%) &  38.65 &  (17.8\%) &  0.9498 &  (28.5\%) & 22\\
No Unprocessing  &   48.28 &  (6.8\%) &  0.9809 &  (7.9\%) &  39.02 &  (14.3\%) &  0.9478 &  (31.2\%) & 22\\
No Unprocessing, 4$\times$ bigger &  48.49 &  (4.5\%) &  0.9818 &  (3.3\%) &  39.35 &  (11.0\%) &  0.9489 &  (29.7\%) & 177 \\
No CCM, WB, Gain  &   48.55 &  (3.8\%) &  0.9817 &  (3.8\%) &  39.70 &  (7.2\%) &  0.9559 &  (18.6\%) & 22\\
Noise-blind  &   48.51 &  (4.2\%) &  0.9816 &  (4.3\%) &  39.81 &  (6.1\%) &  0.9602 &  (9.8\%) & 22\\
No Residual Output  &   48.80 &  (1.0\%) &  0.9824 &  (0.0\%) &  40.19 &  (1.8\%) &  0.9640 &  (0.3\%) & 22\\
No Tone Mapping, Gamma &   48.83 &  (0.7\%) &  0.9823 &  (0.6\%) &  40.23 &  (1.4\%) &  0.9623 &  (4.8\%) & 22\\
